\documentclass[letterpaper, 10 pt, conference]{ieeeconf}  

\IEEEoverridecommandlockouts                              

\usepackage{graphics} 
\usepackage{epsfig} 
\usepackage{mathptmx} 
\usepackage{times} 
\usepackage{amsmath} 
\usepackage{amssymb}  
\usepackage{color}
\usepackage{gensymb}

\newcommand{\myparagraph}[1]{\vspace{0.1in}\noindent\textbf{#1}}

\title{\LARGE \bf
Tactile-Based Insertion for Dense Box-Packing}

\author{
  \authorblockN{Siyuan Dong and Alberto Rodriguez} 
  \authorblockA{
     Massachusetts Institute of Technology\\
    {\tt\small <sydong,albertor>@mit.edu}} 
\thanks{This work was supported by the Amazon Research Awards, and the Toyota Research Institute (TRI). 
This article solely reflects the opinions and conclusions of its authors and not Amazon or Toyota. }
\thanks{We thank Wen Xiong and Rachel Holladay for proofreading the manuscript and Daolin Ma for helpful discussions.}}

\begin{document}

\maketitle
\thispagestyle{empty}
\pagestyle{empty}

\begin{abstract}
We study the problem of using high-resolution tactile sensors to control the insertion of objects in a box-packing scenario.
In this paper, we propose an insertion strategy that leverages tactile sensing to: 
1) safely probe the box with the grasped object while monitoring incipient slip to maintain a stable grasp on the object.
2) estimate and correct for residual position uncertainties to insert the object into a designated gap without disturbing the environment. 

Our proposed methodology is based on two neural networks that estimate the error direction and error magnitude, from a stream of tactile imprints, acquired by two GelSlim fingers, during the insertion process. 
The system is trained on four objects with basic geometric shapes, which we show generalizes to four other common objects.
Based on the estimated positional errors, a heuristic controller iteratively adjusts the position of the object and eventually inserts it successfully without requiring prior knowledge of the geometry of the object. The key insight is that dense tactile feedback contains useful information with respect to the contact interaction between the grasped object and its environment. 
We achieve high success rate and show that unknown objects can be inserted with an average of 6 attempts of the probe-correct loop. The method's ability to generalize to novel objects makes it a good fit for box packing in warehouse automation.
\end{abstract}

\section{Introduction}

Warehouse automation plays an important role in retail for improving efficiency, increasing reliability, and reducing cost. Recently, automatic item picking has experienced an increased interest in the robotics community, due in part to the Amazon Robotics Challenge~\cite{correll2018analysis}.

Automatic item packing is the dual of item picking in warehouse settings. 
Packing items densely improves the storage capacity, decreases the delivery cost and saves packing materials.
It is however a demanding manipulation task which has not been thoroughly explored by the research community. 

\begin{figure}[t]
	\centering
	\includegraphics[width=0.95\linewidth]{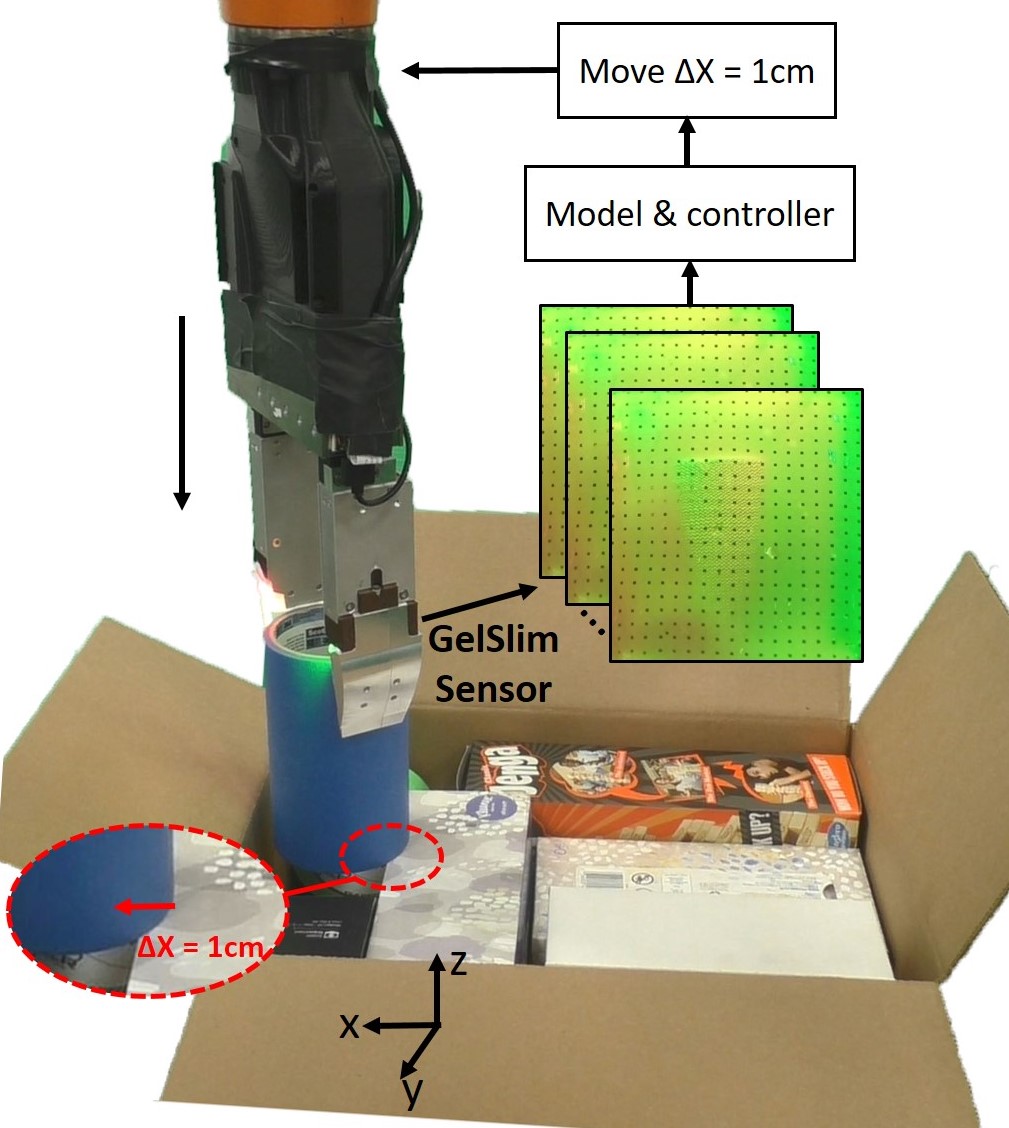}
    \vspace{-4pt}
	\caption{Dense packing task. The robot is inserting a cylindrical tape into a box containing other packed objects. With the position error, the tape is blocked by the surrounding object. The error is detected by GelSlim sensor and passed to the machine learning model and a controller. The controller decides to move to the +x direction by 10 mm, which results in a successful insertion.}
	\label{fig:dense-pack}
\end{figure}

Dense box-packing requires an accurate vision system to estimate the packing position. To avoid collisions, the perception and control systems need to use spatial margins of several centimeters, wasting space. Figure~\ref{fig:dense-pack} shows a collision event in a dense box-packing task.

To address the limitations of vision systems, we propose a packing strategy based on high-resolution tactile information. 
In particular, we use GelSlim~\cite{GelSlim}, a sensor capable of capturing the small changes to the tactile imprints left in the sensor by a grasped object that accidentally collides with the box or surrounding objects. 
The captured images are processed by two neural networks that estimate position errors. Based on these estimated errors, a controller adjusts the insertion position in the next attempt. During the packing process, GelSlim also monitors incipient slip between fingers and object to avoid breaking the grasp. 
We demonstrate that 1) by monitoring incipient slip signals, the system can effectively prevent items from slipping out of the gripper or being crushed by a hard collision, 2) by estimating the relative position between the target object in-hand and the environment objects, small position errors can be corrected, 3) by training with 4 objects with different shapes, the system can generalize to packing new objects. 

Dense packing task is related to the much more explored peg-in-hole insertion problem, but with more tolerance on positional errors and larger object variability. Most solutions to the classic peg-in-hole problem based on either hardware aids or force feedback control~\cite{park2013intuitive,kim1999active} require the knowledge of geometry of the object and the hole, which are not effective for the dense packing task. 

The key insight in this paper is that dense high-resolution tactile information is better suited to the packing problem, than suggested measurements such as with force torque sensors, with which it is more difficult to give a geometric interpretation in terms of contact with the environment without knowledge of the geometry of the object and the environment. 
Our method does not require prior information of the objects and can conform to objects with different geometries. In addition, we only gently and vertically poke the surrounding objects, preventing the surrounding objects from shifting or falling. The tactile-based packing system we propose enables closed-loop control for eliminating position uncertainties.

\section{Related Work}\label{related work}
Dense packing into a cluttered environment has similarities with the peg-in-hole problem. Therefore, in section~\ref{review1} we review different methods for the peg-in-hole problem and discuss whether these methods can be implemented in the dense packing task. Since we adopt a deep learning approach to process image sequences acquired by the vision-based tactile sensor, we review deep learning approaches for image sequence processing in Section~\ref{review2}.

\subsection{Peg-in-hole} \label{review1}
As an active research topic in robotics for decades, the peg-in-hole problem is a typical contact manipulation task that requires precise position and some form of compliance, either passive or active. It represents a large class of assembly tasks in industry. Many approaches have been proposed to solve the problem. One class of methods is based on passive compliance hardware and control algorithms. Drake~\cite{drake1978using} designed a passive compliance device called Remote Center Compliance (RCC) for correcting small uncertainties in the assembly task. Whitney~\cite{whitney1982quasi} further analyzed deformations of the geometry and the forces of rigid part mating. The parameters of the RCC device were tuned to adjust the peg orientation relative to the hole. Jain~\textit{et al}.~\cite{jain2013scara} demonstrated that adding compliance with two ionic polymer metal composite (IPMC) compliant fingers had advantages in peg-in-hole assembly. Park~\textit{et al}.~\cite{park2013intuitive} proposed a strategy that adopted hybrid force/position control and passive compliance control for successful peg-in-hole assembly. 

Another class of methods is model-based active sensing. By utilizing feedback from sensors to identify the configurations or errors in the assembly process, these methods are usually more adaptable to new environments. Bruyninckx~\textit{et al}.~\cite{bruyninckx1995peg} proposed a model-based method to model different contact situations and deployed the model and feedback from a force sensor to explicitly find the hole and align the axes of the peg with the hole. Kim~\cite{kim1999active} developed an insertion algorithm according to the quasi-static analysis of normal and tilted modes with force/moment sensors. These model-based methods leveraged the geometry of the peg, which was usually cylindrical. These methods are ineffective for the packing task, where object geometry varies and is often not known. 

Learning-based methods instead utilize patterns of data acquired by the sensor in different situations to guide the hole search and motion corrections. These sensors are most often force torque sensors. Newman~\textit{et al}.~\cite{newman2001interpretation} proposed a method to interpret the force and the momentum signal in a preliminary assembly attempt, and used it to find holes in the compliant peg-in-hole task. Gullapalli~\textit{et al}.~\cite{gullapalli1994learning} trained a policy that learned the mapping between the contact force resulting from misalignment and the motion that reduced the misalignment. As the computation power increases, more and more vision sensors are used to solve the peg-in-hole problem. Levine~\textit{et al}.~\cite{levine2016end} trained an end-to-end strategy which took images from an external camera as the input and directly outputted applied robot motor torques. They demonstrated the policy in several peg-in-hole like assembly tasks. Lee~\textit{et al}.~\cite{lee2018making} proposed a deep reinforcement learning method with both images and forces as the input. They used self-supervised learning to learn the representation of the images and the forces in advance, which improved the data efficiency of the learning process. In addition, the policy can be generalized to several pegs with different shapes. 

Our method belongs to the learning-based active sensing methods. Different from the peg-in-hold problem, the surrounding objects in the dense packing task are not fastened. Therefore the searching strategies with force control~\cite{park2013intuitive,kim1999active,chhatpar2001search} that rely on an known geometry are not appropriate. To retain the position of surrounding objects, our method performs a vertical poking to the surrounding objects, instead of sliding into them. It's worth pointing out that the GelSlim sensor uses a soft gel layer as the contact surface, yielding some compliance to facilitate the insertion task.
\subsection{Deep Learning for Image Sequence Processing}\label{review2}

Deep learning approaches for processing image sequences are mostly used in action classification tasks in computer vision. According to whether 2D or 3D convolutional kernels are deployed, the neural networks can be classified into two main classes. The ConvNet+LSTM~\cite{donahue2015long} method we use here has 2D kernels. It employs CNN to extract spatial features and LSTM to capture temporal features. This method can benefit from classical CNN models pretrained with ImageNet. Carreira~\textit{et al}.~\cite{carreira2017quo} proposed the state of the art 3D convolution method, which is basically an inflated Inception net~\cite{inception_net}. It requires not only the raw RGB image sequence but also the corresponding optical images at the input. This type of two-stream network fusion style can be naturally adapted to our task, because we can obtain the optical images easily by tracking the markers on GelSlim sensor.

There are also related works in robotics that extract useful information from sequential image signals with deep learning. Nguyen~\textit{et al}.~\cite{nguyen2018translating} proposed a new method to translate videos to commands in robotic manipulations, enabling robots to perform manipulation tasks by watching human behaviors. Finn~\textit{et al}.~\cite{finn2016unsupervised} developed a video prediction model that predicted the pixel motion (object motion). Based on the actions and the initial frames of the robot, the model predicts the next several frames in a robot pushing task. The architecture of the model belongs to the family of CNN+LSTM. Lee~\textit{et al}.~\cite{lee2017learning} designed a convolutional and temporal model that allowed robots to learn new activities from unlabeled example videos of humans. The basic idea was to learn the temporal structure of human activity and then apply to the motor executions of robots. 

Several prior works use video signals from vision-based tactile sensors to extract physical properties or motion information. Yuan~\textit{et al}.~\cite{yuan2017shape} demonstrated that the GelSight tactile sensor could estimate the hardness of objects, since the sensor generated different image sequences when contacting with soft and hard objects. They trained a CNN+LSTM model to learn the hardness of objects directly from the image sequences in the contact period. They also applied a similar principle to infer the physical properties of different clothes~\cite{yuan2018active}. Li~\textit{et al}.~\cite{li2018slip} trained a neural network with image sequences from  GelSight and an external camera to detect slip. Zhang~\textit{et al}.~\cite{zhang2018fingervision} performed an analogous experiment with a different tactile sensor. The slip detection task with tactile sensors is similar to our task to some extent, because both require tracking local motion of the object in hand.

\section{Box-Packing TASK}
\myparagraph{Task Description} We perform the dense packing task under position uncertainties. We assume the position of the gap has been roughly estimated by a vision system and the initial position of the object to pack is known. The packing environment is shown in Fig.~\ref{fig:set-up}. We introduce controlled position errors in x direction (translation) and in yaw (rotation). The range of translation errors and rotation errors are $-30\%\sim30\%$ of the object's width and $-15\degree\sim-15\degree$, respectively. The assembly clearance is about 2 mm. The two objects on the side of the hole are not fixed in test experiments. We assume no prior knowledge of the geometry of the target object. We perform the task on a series of objects with different shapes.

\myparagraph{Performance Criteria} We ask the insertion algorithm three goals: 1) the target object is firmly grasped in the gripper in the insertion process. 2) the robot does not change the positions of the surrounding two objects. 3) the robot is able to correct the position error within several trials for realistic applications.

\section{Method}
Humans can roughly estimate a correction signal in a blind insertion process, especially if the positional error is relatively small. 
The tactile sensors in our compliant fingers can detect the small motion of the object in-hand when the in-hand object collides with the surrounding objects. The key idea we exploit is that the tactile signal generated during that contact event is correlated with the contact formation, and the relative position and orientation between object and gap. 

\begin{figure}[t]
	\centering
	\includegraphics[width=0.95\linewidth]{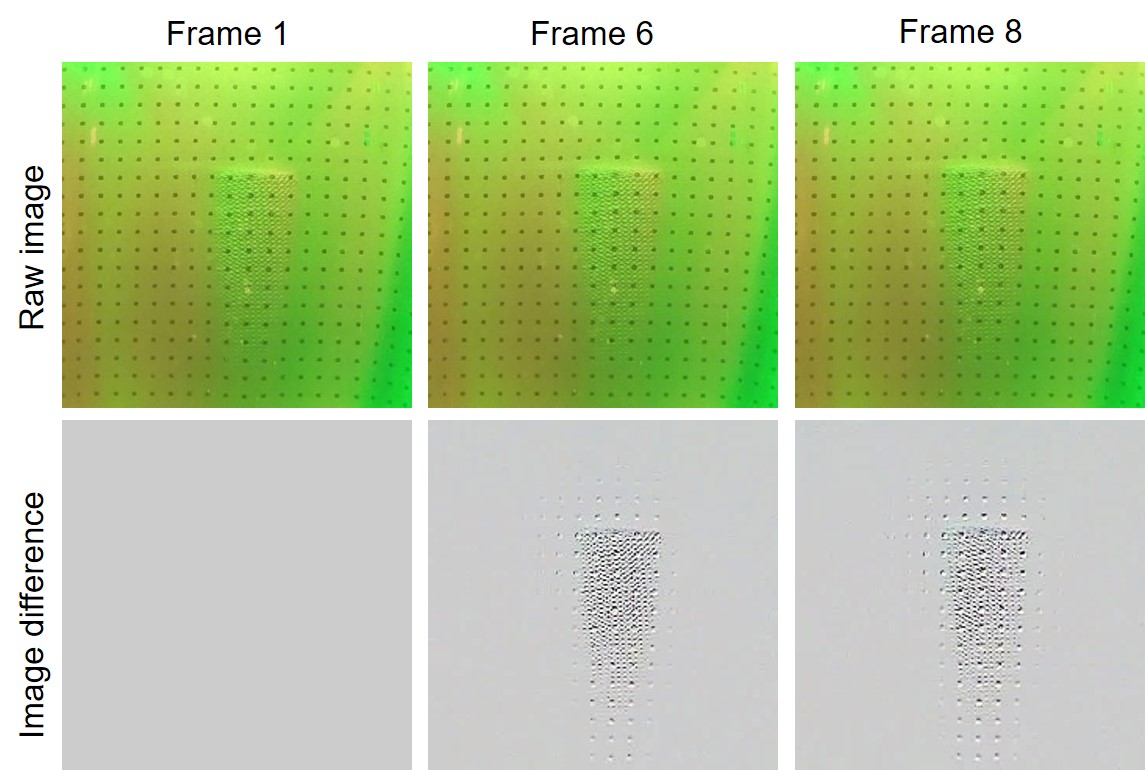}
    \vspace{-6pt}
	\caption{Example image sequence of a cylindrical object in the collision process, captured by the GelSlim sensor. Top rows: raw images (frame 1, 6 and 8). Bottom row: differences between the corresponding raw image and frame 1.}
	\label{fig:data_sample}
\end{figure}

\myparagraph{GelSlim sensor} 
To detect the motion of the object during contact, we use the GelSlim~\cite{GelSlim} vision-based tactile sensor (Fig.~\ref{fig:dense-pack}). The contact surface of the sensor is a piece of soft elastomeric gel, covered with a textured wear-resistant cloth. Black dots that move with applied shear force are uniformly marked on the gel. A camera captures the deformation of the gel surface, as well as the motion of the markers.
Force/strain information can be estimated from the motion of the markers on the gel surface~\cite{ma2018dense}. The soft sensor surface exhibits compliance, enabling the sensor to detect the 3D motion of the object in-hand and slip~\cite{dong2018slip}. The motion of the object during contact is encoded in the sequence of images captured by the sensor, shown in Fig.~\ref{fig:data_sample}. To estimate the relative position of the object and the hole from the image sequence, we use a deep learning network.

\begin{figure}[t]
	\centering
	\includegraphics[width=1.0\linewidth]{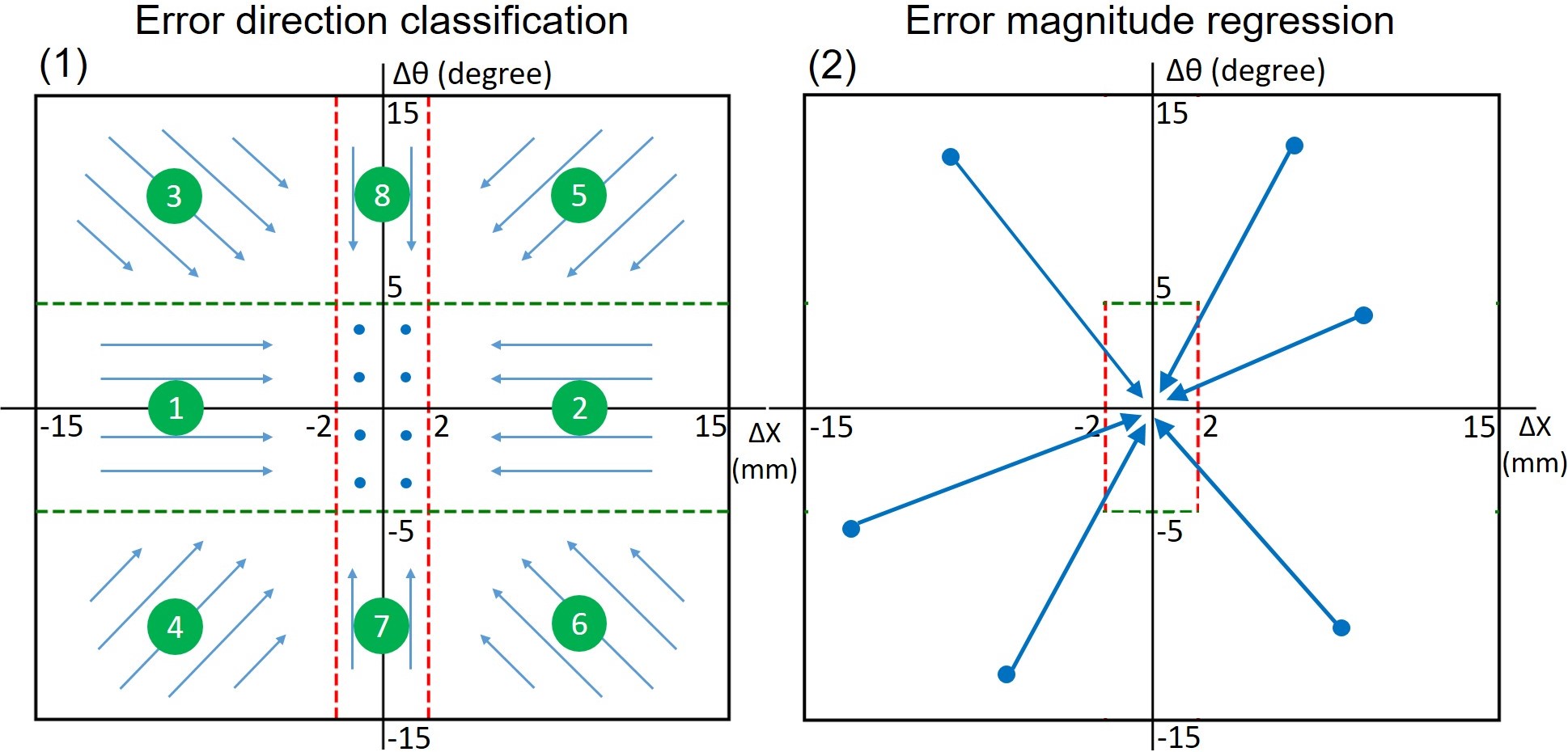}
    \vspace{-20pt}
	\caption{(1) 8 regions categorized by the error direction in the error space. (2) Schematic illustration of the error magnitude estimation. The length of the arrow represents the magnitude of the error. The direction of the arrow aligns with the error direction. }
	\label{fig:error classification}
\end{figure}

\myparagraph{Classification of error directions} Yu~\textit{et al.}~\cite{yu2018realtime} categorized the contact formation of a rectangular prism with a parallel gap as in our case, into 8 classes, according to which edge of the object was contacted with the environment. Because we are aimed at generalizing the method to objects with different shapes and estimating the direction of the error, here we categorize the contact situations into 8 classes according to different combinations of the directions of two errors (x and $\theta$). We set $T_{\theta} = 5\degree$ as the threshold of rotation errors and $T_x = 2.5$ mm as the threshold of translation errors in our experiment, as shown by the red and green dash lines in Fig.~\ref{fig:error classification}. For example, the region in the upper left corner (class 3) represents the translation error along $-x$ direction and the rotation error along $+\theta$ direction. The neighboring region on the right refers to only the rotation error along $+\theta$ direction. 
Our hypothesis is that different error directions will result in distinguishable tactile imprints.

When the object held by two GelSlim fingers collides with the surrounding objects, it rotates along the edge of the environment object. If there is only a translation error, the rotation of the object is parallel to the gel surface, resulting in marker motions on the sensor surface. However, if a rotation error occurs, the rotation direction of the object can be decomposed into the directions parallel and perpendicular to the gel surface. It results in both marker motions and normal pressure changes on the gel surface. This fact lays the foundation of distinguishing error directions from the image sequences with a neural network. Furthermore, after obtaining the error signal, we can reduce the error by moving the object in the direction that negates the error gradually. Since the output of the classifier are the probabilities of each class, it interpreted the confidence of the neural network in the current state. We call this neural network \texttt{DirectionNN}.

\myparagraph{Regression of error magnitude} 
Inserting the object into the hole with a minimal amount of trials requires more than the direction of the error. We experimentally observe that errors with different magnitudes but same direction also produce distinctive image sequences. Therefore, we train another neural network to estimate the error magnitude. Regression with a neural network is more difficult than classification, but even a rough estimate of the error is helpful for efficiency. We name this neural network \texttt{MagnitudeNN}. %
We combine the outputs of two independent models, which can boost the performance and robustness of the model. 

\begin{figure}[t]
	\centering
	\includegraphics[width=\linewidth]{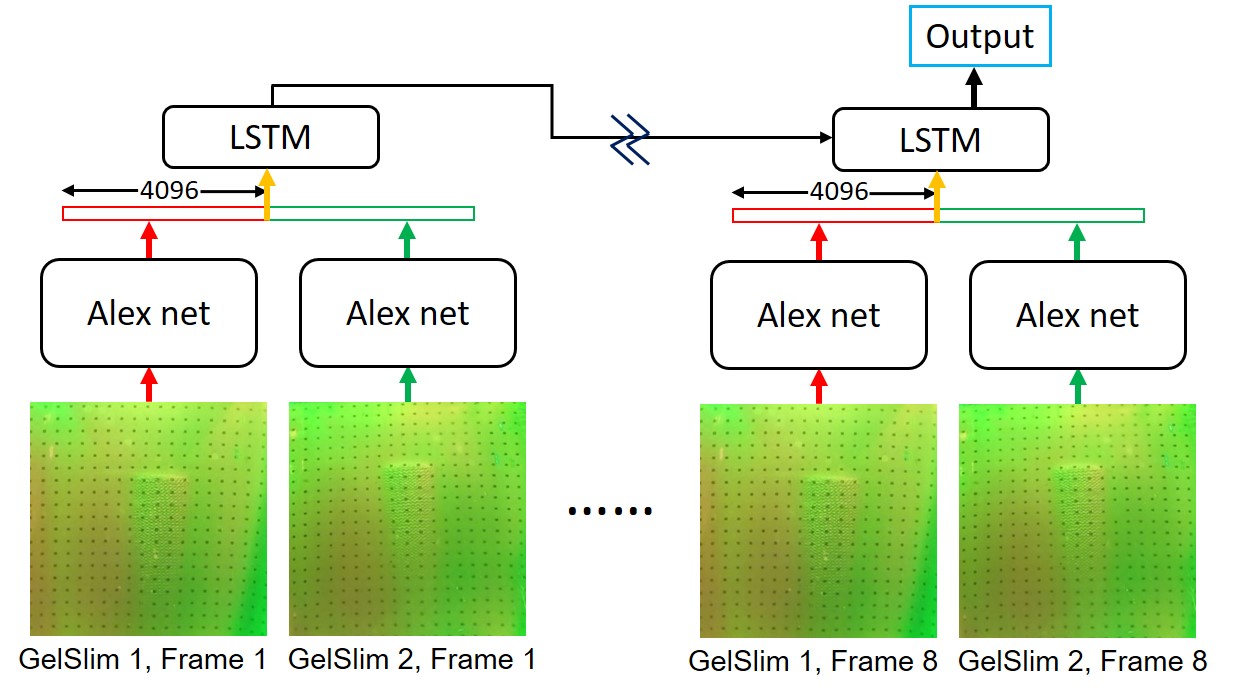}
    \vspace{-20pt}
	\caption{Architecture of the CNN+LSTM neural network. The image sequences from two GelSlim sensors are processed by the Alexnet and then concatenated. The output features from Alexnet feed the LSTM for predicting the errors of the insertion position.}
	\label{fig:Architecture} 
\end{figure}

\myparagraph{Model} As discussed above, we train two independent neural networks to classify the error directions and to regress the error magnitude, using a Convolutional Neural Networks (CNN) + Recurrent Neural Networks (RNN) architecture (Fig.~\ref{fig:Architecture}). The CNN with Alexnet~\cite{Alexnet} architecture (the last layer removed) extracts 4096 features from each image. The input data is two streams of images from two GelSlim sensors. Because the contact period is very short, we select 8 frames starting from the frame that first captures the motion of the object. We pair the 16 images into 8 pairs by the sequence order. The two lists of features from each pair of images are concatenated into a list of 8192 features. Afterward, the 8 lists of features are processed by an RNN model (LSTM)~\cite{LSTM} with 1 hidden layer and 170 hidden units. The output of \texttt{DirectionNN} is the number of regions in error space, and the output of \texttt{MagnitudeNN} is estimated errors in $x$ ($\Delta x_e$) and $\theta$ ($\Delta \theta_e$). Though the architectures of the two networks are the same, they are trained independently.   
\begin{figure*}[t]
	\centering
	\includegraphics[width=0.8\linewidth]{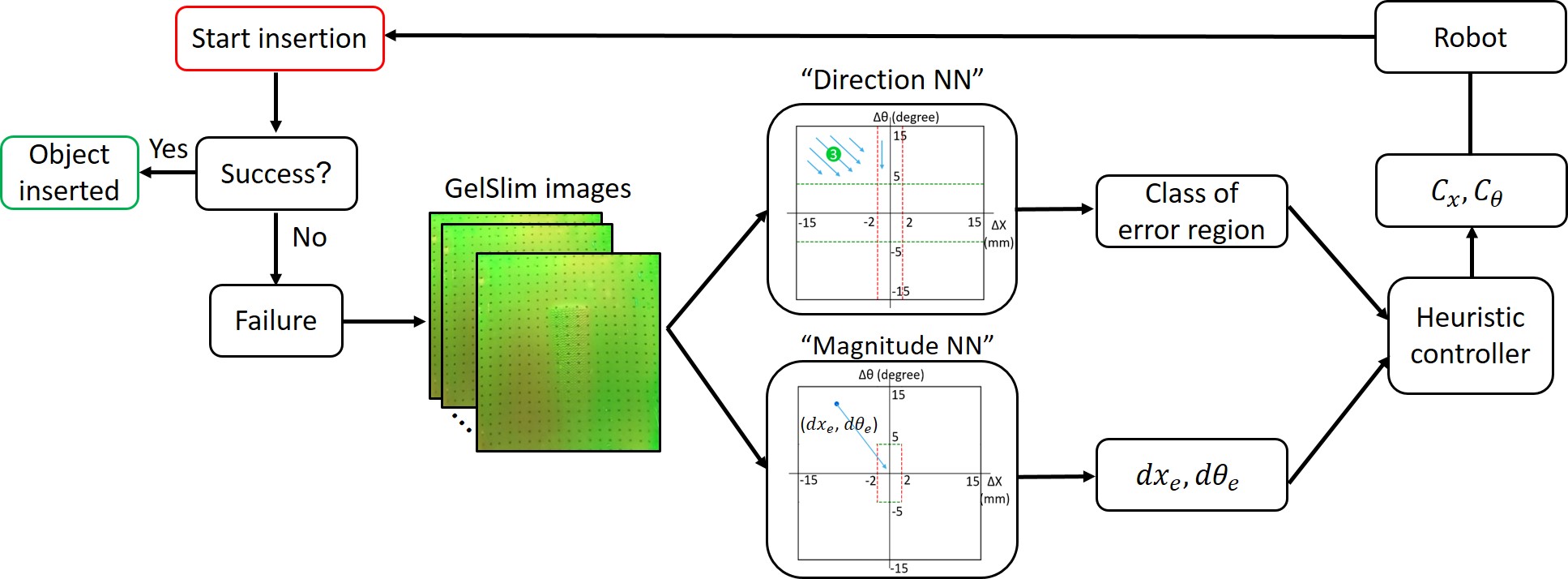}
    \vspace{-4pt}
	\caption{Workflow of our control strategy. The robot starts the insertion and if successful, the process is done. If the object collides with the environment, the motion of the object in the contact period is captured by the image sequences of GelSlim sensors. The image sequences are used to train two neural networks. The class of error is estimated by \texttt{DirectionNN} and the magnitude of the error is predicted by \texttt{MagnitudeNN}. The output from the neural networks is passed to a heuristic controller. The controller decides adjustment of the insertion position. If at the new position the insertion is successful, the process ends. Otherwise, the new error is again detected by GelSlim sensors. The loops run until success.}
	\label{fig:flowchart} 
\end{figure*}

\myparagraph{Control strategy}
Based on the estimation of the error magnitude and direction, we propose an heuristic controller to correct the position error. The class number obtained from the \texttt{DirectionNN} is converted to the signs of the errors in x ($S_x$) and $\theta$ ($S_\theta$). $S_x$ and $S_\theta$ can take a value of 1, 0, or -1 with -1 and 1 indicating errors in the positive or negative direction and 0 indicating no error. The outputs of the controller are $C_x$ and $C_\theta$, which are the corrections in x and $\theta$ directions. 

We formulate the control strategy with Eq.~\ref{eq:1}. When the estimations from the two models are consistent, the corrections are directly -$\Delta x_e$ and -$\Delta \theta_e$, shrunken by a factor 0.7, which experimentally helps avoid overshooting. When the \texttt{DirectionNN} indicates no error in the direction, we further decrease the factor to 0.3. We observe from experiments that estimating the error direction is easier and more robust than estimating the error magnitude. Therefore, we trust the \texttt{DirectionNN} prediction if the two models are contradictory. In this case, a constant step (3 mm or 3\degree) is chosen for $C_x$ and $C_\theta$. To prevent overshooting after one trial, we clip the magnitude of $C_x$ and $C_\theta$ at 4 mm and 4\degree for later trials. We summarize the workflow of the box-packing approach in Figure~\ref{fig:flowchart}.

\begin{equation}
  \label{eq:1}
  \begin{gathered}
    C_x =    
    \begin{cases}
    -0.7\times \Delta x_e      & \text{if $S_x \times \Delta x_e>0$} \\
    -0.3\times \Delta x_e      & \text{if $S_x=0$} \\\
    -3.0\times S_x & \text{if $S_x\times \Delta x_e<0$}
    \end{cases}\\    
    C_\theta=    
    \begin{cases}
    -0.7\times \Delta  \theta_e  & \text{if $S_\theta \times \Delta \theta_e>0$} \\
    -0.3\times \Delta  \theta_e & \text{if $S_\theta=0$} \\\
    -3.0\times S_\theta & \text{if $S_\theta\times \Delta \theta_e<0$}
    \end{cases}\\
  \end{gathered}
\end{equation}

\section{Experiments}
In this section, we explain the details of the experimental setup, data collection process, parameter tuning for model training and the test experiments for evaluating the system.
\subsection{Experimental setup}
\begin{figure}[t]
	\centering
	\includegraphics[width=0.9\linewidth]{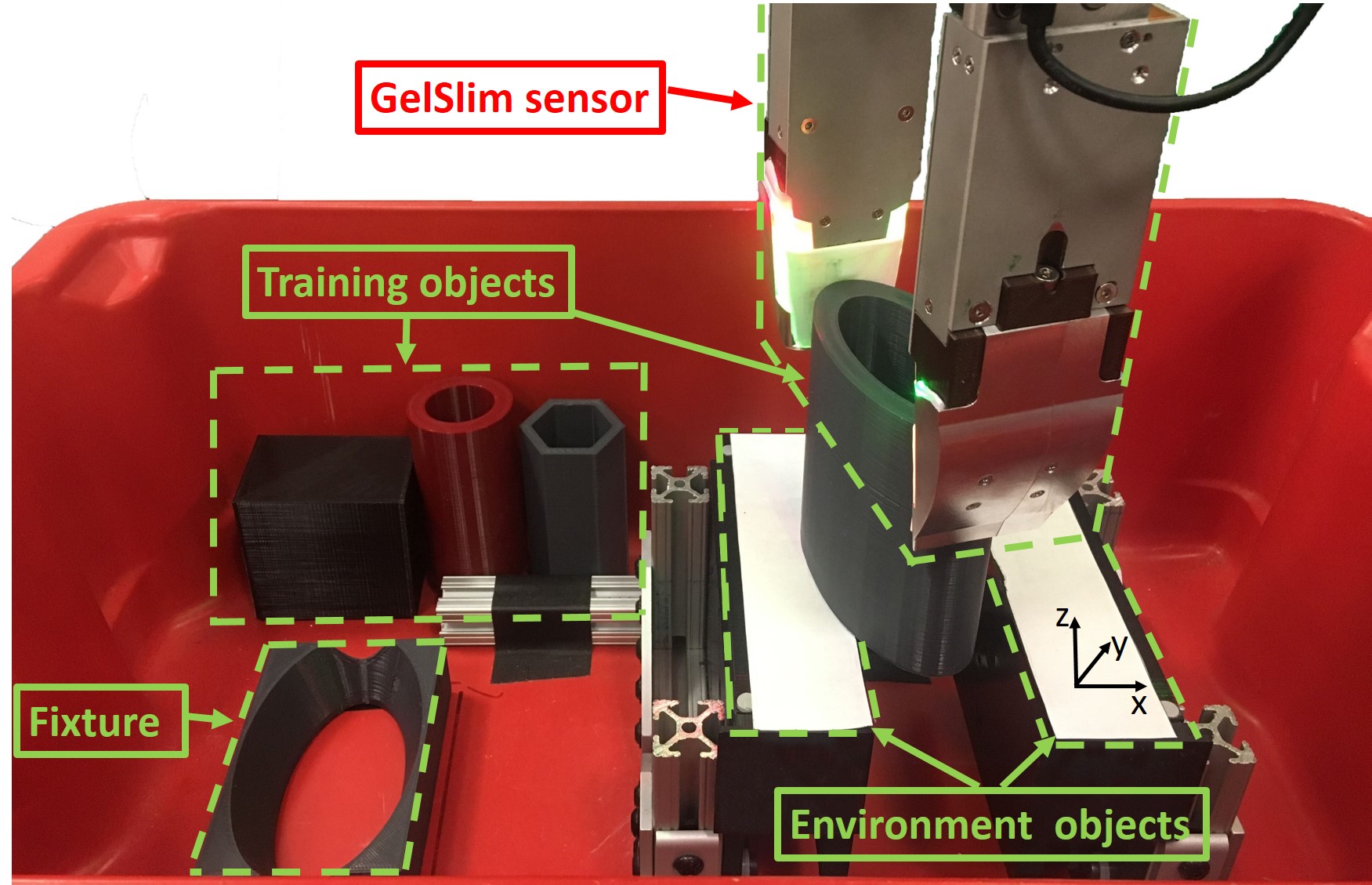}
    \vspace{-6pt}
	\caption{Experimental setup. The system is composed of two GelSlim sensors, 4 fixtures (only one is shown) for adjusting the pose of the object in the gripper, two environment objects and 4 objects for training the neural networks.}
	\label{fig:set-up}
\end{figure}

The experimental setup shown in Fig.~\ref{fig:set-up} is a simplified version of the grasping system developed in~\cite{Zeng2018}. The system includes a 6-DOF ABB-1600 robot arm, two GelSlim sensors attached to a WSG-50 parallel gripper. The packing environment is designed as two 3D printed rigid blocks (top surface: $45\times155$ mm) with 56 mm gap between them. To ensure that the data labelling is accurate, the two environment objects are fixed in the data collection process. In the test experiments, the two objects are free to move. 

We 3D print 4 target objects of different shapes. The base shapes are a circle ($\O=25.5$ mm), a rectangle ($51\times80$ mm) , an ellipse ($a=105$ mm, $b=25.5$ mm) and a hexagon ($R=25.5$ mm). We prepare a fixture to precisely relocate each object after a few experiments. Fig.~\ref{fig:set-up} shows the fixture for the ellipse object.

\subsection{Data Collection}\label{data collection}
We use a self-supervised scheme to collect and label the packing data for a set of training objects under controlled positional errors. We first sample translation and rotation errors uniformly from $-15$ mm $< x < 15$ mm, $-15\degree < \theta < 15\degree$. As the position of the gap is known, we can introduce controlled error by adding the position error to the gap position. The robot moves down vertically while the slip detection function~\cite{dong2018slip} of GelSlim sensor monitors slip. If the slip detection is triggered earlier than expected, the object is blocked by the surrounding objects. In this case, we log the last 30 frames of GelSlim images and the corresponding position error. Otherwise, the robot inserts the object successfully, and we do not need to log the data. The robot repeats the process described above with different errors until enough data is acquired. To better observe the motion of the object, we set the grasping force slightly higher than the minimum force required. The pose of the object in hand changes after several pokes. To acquire consistent data, the robot puts the object back to the fixture to correct the pose after every 5 trials. We collect about 7500 data for each object and about 30000 data in total.  

\subsection{Model Training} 
\myparagraph{Data preprocess} Since circle and hexagon objects do not have class 7 and 8 errors (only rotation errors and no translation errors), the data becomes unbalanced. We manually double the entire data of class 7 and 8 with data augmentations (random cropping). Because Alexnet is pretrained for object recognition, the absolute location of the tactile imprints in the sensor image is neglected. However, the relative position information for us is crucial. Therefore, instead of using the raw images, we use image differences as the input data (example in Fig.~\ref{fig:data_sample} second row)

\myparagraph{Training specifics} For faster training, we freeze the convolutional layers of Alexnet and train a smaller neural network with 2 fully connected layers of CNN and LSTM. We use ADAM optimizer~\cite{kingma2014adam}, NVIDIA Titan X Pascal GPU and Pytorch package for training. 

\subsection{Test experiment}
\myparagraph{Test with training objects} We first test the system with the 4 training objects to evaluate whether the neural networks can learn useful features. The two environment objects with a 56 mm gap are released and free to move or fall during test. We test 26 packing attempts with random positional errors. To test the method in the extreme error conditions, we also test the four extreme error conditions where $\Delta x$ and $\Delta \theta$ saturate in both positive and negative directions. 

In the experiment, the robot picks the target object from a known position and moved it to a noisy position above the gap. It vertically inserts the object and stops if incipient slip is detected. If the object is blocked by the environment objects, the GelSlims signals during the contact period will be fed to the two neural networks and then the estimation of the errors from the neural networks will be sent to the controller. The robot adjusts the pose according to the $C_x$ and $C_\theta$ produced by the controller and start the next trial. The process repeats until the object is successfully inserted or it stops after 15 trials. We record the number of trials and perturbations to surrounding objects to evaluate the performance.

\myparagraph{Test with new objects} We further test the system with 4 daily objects that the neural networks have never seen. The objects (Table~\ref{tab:objects}) have different base shapes, weights, and widths. The vitamin bottle is cylindrical, similar to the cylinder object in the training. However, the bottle cap part the robot grasps in the experiment has sharp textures. The radius is also smaller than that of the base. The white box is a rectangle but with different sizes from the box in the training. Box packing or stacking is very common in warehouse, so it is important that the method works for boxes with different sizes. The mustard container and the metal can are heavy and their base shapes are rounded rectangles with different curvatures. They all have flat bottoms because the method we propose only considers the contact problem in 2D space. Taking into account the different sizes of the 4 daily objects, we change the width of the gap to keep the clearance to be 2mm when testing each object. The range of the translation error x is also adapted to the specific object, from -30\% to 30\% of the width of the object. The rest of the experiment is the same as the experiment of training objects described above.

\section{Results}
\subsection{Model accuracy}
\begin{figure}[h]
	\centering
	\includegraphics[width=1.\linewidth]{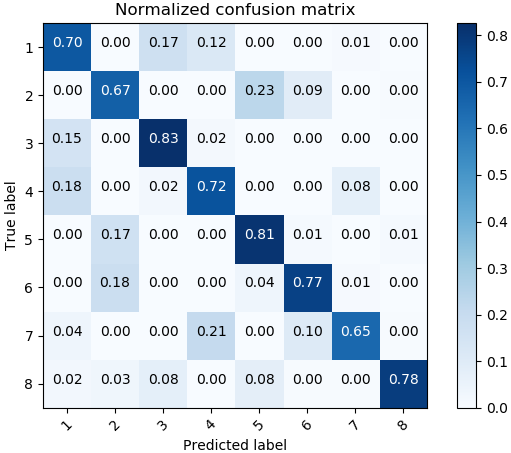}
    \vspace{-4pt}
	\caption{Confusion matrix of the \texttt{DirectionNN} for training objects. The diagonal elements represent the classification accuracy for each class. The off-diagonal elements are the error rates of classifications to a wrong class.}
	\label{fig:confusion_matrix}
\end{figure}
The training accuracy of the \texttt{DirectionNN} is 80.5\% and validation accuracy is 74.4\%. The relatively similar values of training and validation accuracies indicate that the system does not overfit. The confusion matrix in Fig.~\ref{fig:confusion_matrix} demonstrates the effectiveness of the trained classifier. The large diagonal values confirm the high accuracy. The off-diagonal elements also give hints about the types of misclassification. For example, the top row indicates that the model is confused by class 3 and 4 when the truth is class 1. It makes sense that they all have the -$x$ error, lying on the left-hand side of the error space  (see Fig.~\ref{fig:error classification}). Similarly, the bottom row shows that it is more difficult for the model to distinguish class 8 from class 3 and class 5, because they all have errors in +$\theta$ direction. The model can always distinguish -x from +x directions and it is similar for the rotation errors. By following the classifier, a simple controller that moves small steps along the opposite direction can highly probably insert the object into the designated gap.
\begin{table*}[] 
\centering
\vspace{6pt}
\caption{Success rate, mean and max number of trials of the tests for 4 training objects and 4 new objects} \label{tab:objects} 
\begin{tabular}{c|c|c|c|c|c|c|c|c}
\hline   
  &  Rectangle & Circle &  Ellipse & Hexagon & White box & Vitamin bottle & Metal can & Mustard container \\ \hline
   & \includegraphics[width=0.062\textwidth]{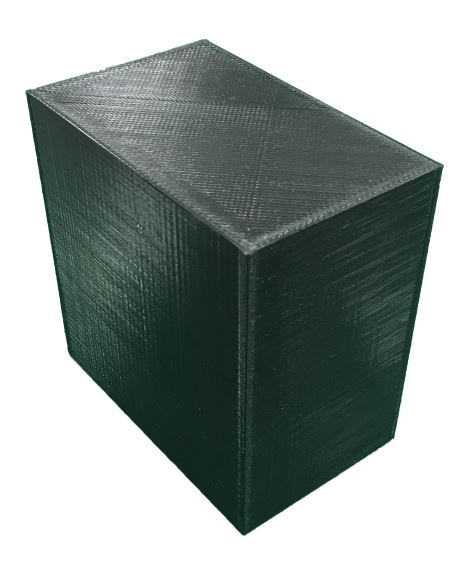}  & \includegraphics[width=0.05\textwidth]{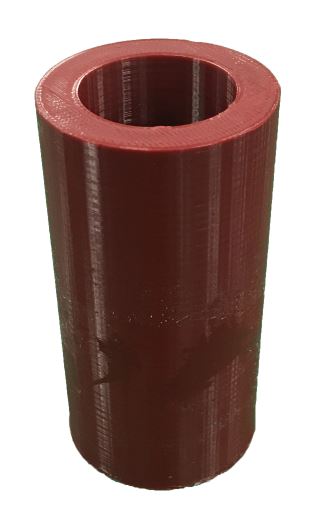}  & \includegraphics[width=0.055\textwidth]{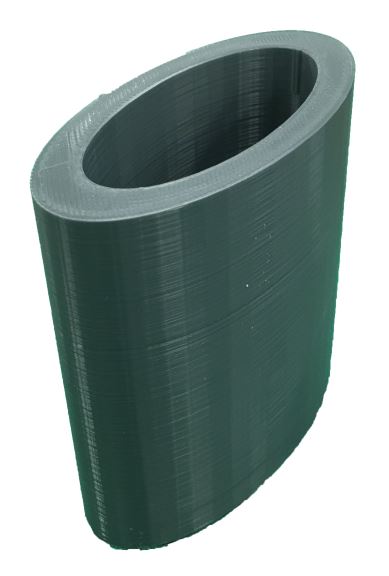}  & \includegraphics[width=0.048\textwidth]{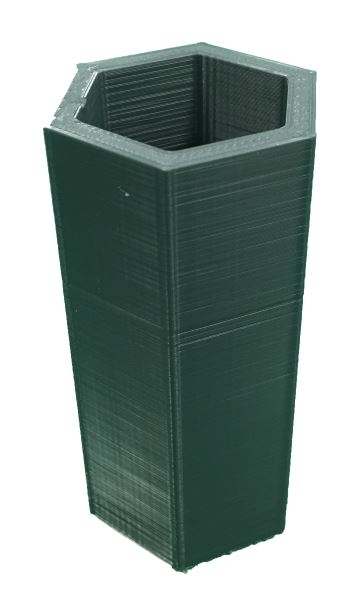}  & \includegraphics[width=0.065\textwidth]{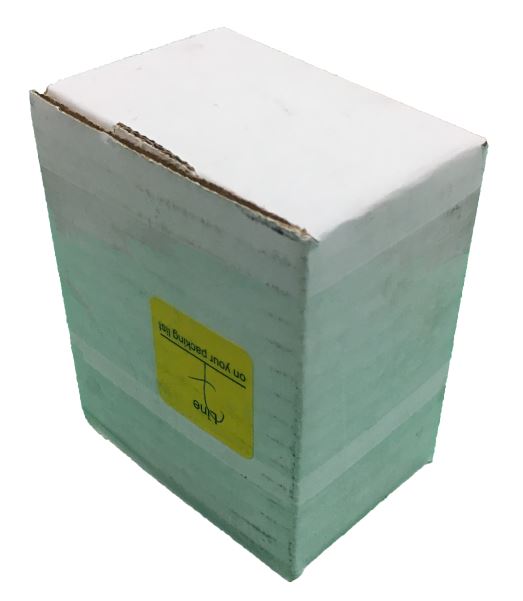}  & \includegraphics[width=0.043\textwidth]{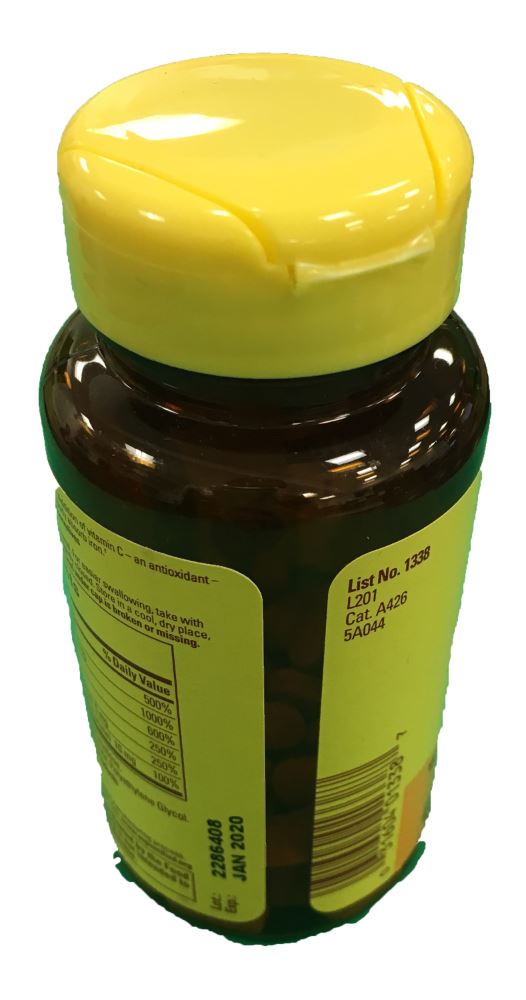}  & \includegraphics[width=0.067\textwidth]{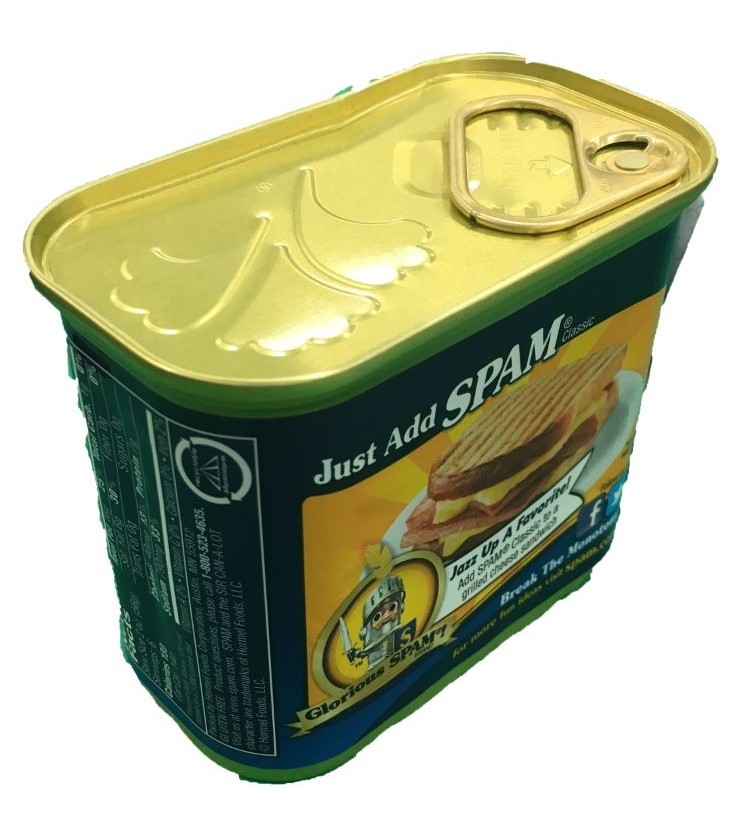} & \includegraphics[width=0.045\textwidth]{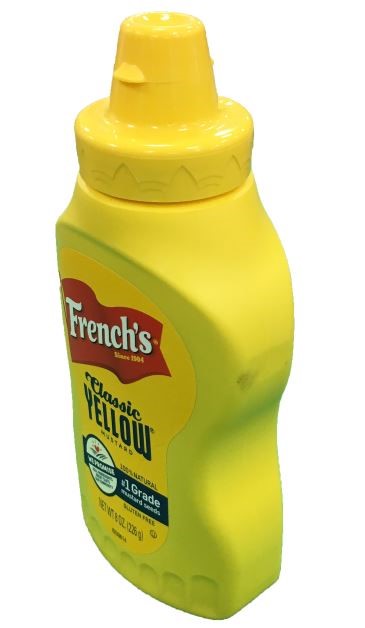}  \\ \hline
 Success Rate & 100.0\%  & 100 \% & 100.0\% & 100.0\% & 93.3\% & 96.7\% & 93.3\% & 100\% \\ \hline 
 Mean number of trials & 4.7 & 2.97  & 2.77 & 2.4 & 6.53 & 2.97 & 7.4 & 3.6 \\ \hline 
 Max number of trials & 11 & 5 & 8  & 6 & $>15$ & $>15$ & $>15$ & 9 \\ \hline 
\end{tabular}
\end{table*}
The training loss of the \texttt{MagnitudeNN} is 10.5 and the validation loss is 14.4. The averaged error for translation error $\Delta x$ and rotation error $\Delta\theta$ are estimated to be -1.9 mm$\sim$1.9 mm and $-1.9\degree\sim1.9\degree$, assuming the two errors contribute equally to the mean square loss. Though the error magnitude estimation performs better than expected, we still use a discount factor in the controller to avoid overshooting.  

\subsection{Experimental results with training objects }
\begin{figure}[t]
	\centering
    \includegraphics[width=1.0\linewidth]{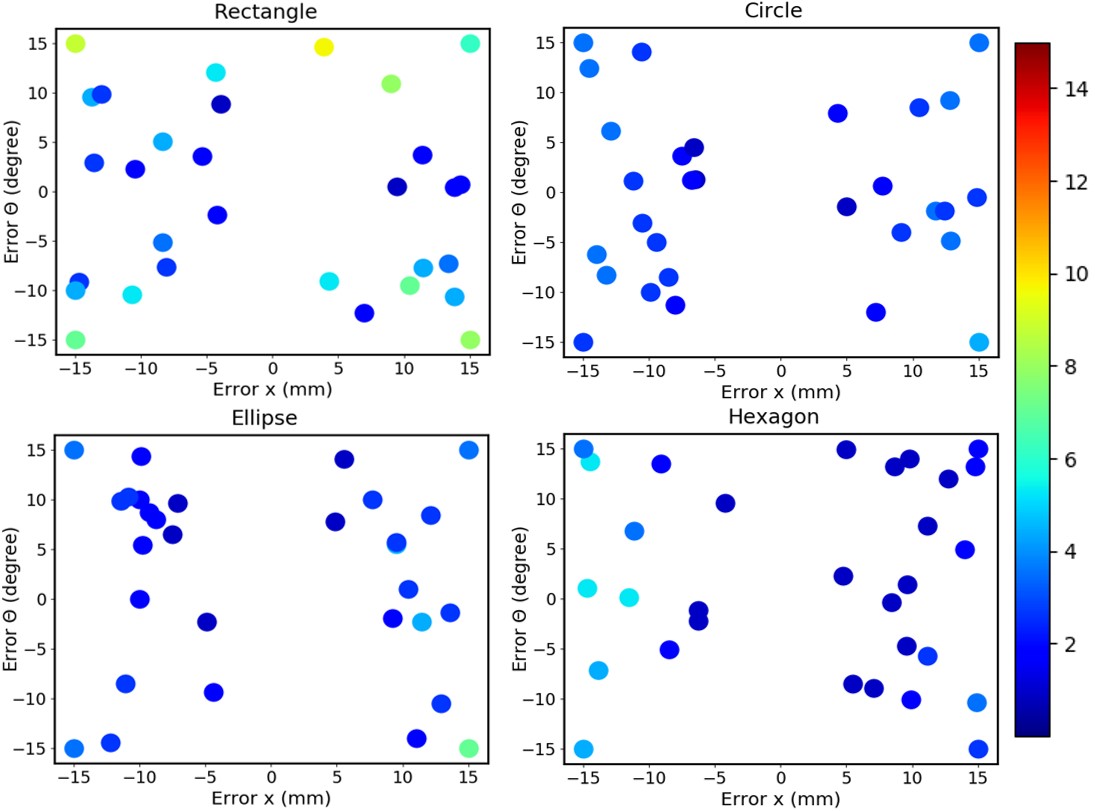}
    \vspace{-16pt}
	\caption{Number of trials before a successful insertion for the 4 training objects with different initial position errors. The rotation error $\theta$ changes from -15$\degree$ to 15 $\degree$ and the translation error $x$ varies from -15 mm to 15 mm. The color of the points corresponds to the number of trials.}
	\label{fig:result_train}
\end{figure}

In 120 tests of the 4 training objects, the objects were inserted successfully 100\% within 15 trials. The two surrounding objects were stable during the test. We plot the number of trials vs. error $x$ and error $\theta$ for each test in Fig.~\ref{fig:result_train}. The colors of the points represent the number of trials. We summarize the successful rate, mean and max number of trials for each object in Table~\ref{tab:objects}. 

The rectangle object is the most difficult case, because even a slight rotation error will result in a collision. The average number of trials of inserting the rectangle object is 4.7, twice higher than that of the other 3 objects. In addition, from Fig.~\ref{fig:result_train}, almost all tests that require more trials fall in the region of large rotation error $\theta$. It is due to the fact that a large rotation error reduces the object motion parallel with the sensor plane that is easy to capture, and increases the motion perpendicular to the sensor surface that is hard to capture. 

These results demonstrate that the two neural networks and the simple control strategy perform well on the 4 training objects. The 2D textures of the 4 objects are very different, but the neural network can extract useful information for the insertion task. Though the accuracy of a single neural network is not high, the complimentary contributions of the two neural networks and the control strategy significantly benefit the decision making. By estimating the error magnitude, \texttt{MagnitudeNN} reduces the number of trials when the error is large, as shown by the large error regions in Fig.~\ref{fig:result_train}. 

\subsection{Experiment results with new objects }
\begin{figure}[h]
	\centering
    \includegraphics[width=1.0\linewidth]{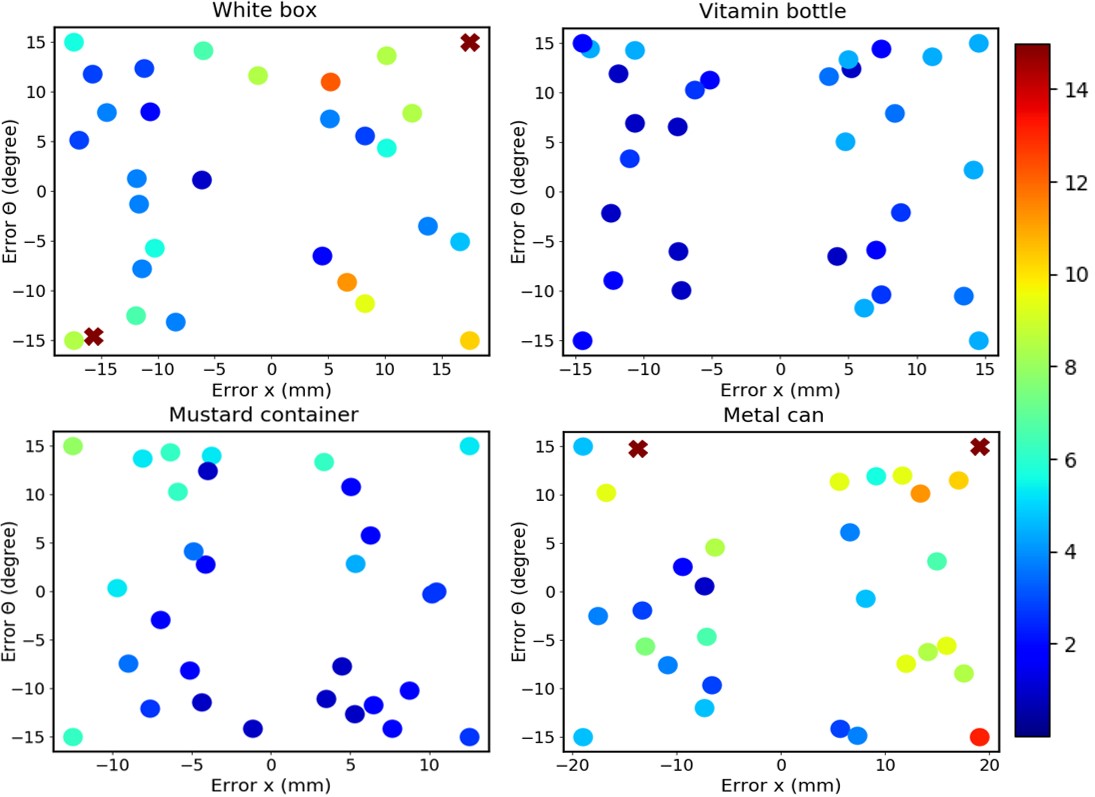}
    \vspace{-16pt}
	\caption{ Number of trials before a successful insertion for the 4 new objects with different initial position errors. The rotation error $\theta$ changes from -15$\degree$ to 15 $\degree$ and the translation error $x$ varies from -15 mm to 15 mm. The color of the points corresponds to the number of trials. The dark red crosses indicate failed insertions after 15 trials.}
	\label{fig:results_new}
\end{figure}
We also achieved a high successful rate for the 4 new objects. In 30 tests of each object, the mustard container has 100\% success rate. There is only 1 failure case for the Vitamin bottle and 2 failure cases for the other 2 objects. We plot the number of trials at each error point in Fig.~\ref{fig:results_new} and summarize the successful rate, mean and max number of trials in Tabel~\ref{tab:objects}. The dark red crosses in Fig.~\ref{fig:results_new} represent the failure cases, which always happen in the region of large rotation or translation errors. The number of trials of the failures is set to 16 for calculating the mean.  

The vitamin bottle and the mustard container both have rounded edges and require only a few trials (averagely 2.97 and 3.6). The numbers are similar to the average numbers of trials of the circle, ellipse and hexagon objects, confirming the adaptability of the neural networks to new objects with rounded base shapes. The base shapes of the metal can and the white box are close to a rectangle. Comparing to the mean number of trials of the trained rectangle object, the robot needs 2 or 3 more trials to insert the two new objects. If we remove the 2 failure cases, they only require one more trial than rounded shapes. All of the failure cases happened under bigger rotation error and translation error. An incorrect estimation of the error direction in the first trial makes the controller diverge and it takes several attempts to recover and capture useful information to locate the gap.

Overall, our method can be directly generalized to the insertion task of objects with rounded base shape. For objects with rectangle base shape, our method performs a little bit worse than the result of the trained object. Considering all tested new objects, it takes averagely 6 trials to insert the object into the gap successfully.

\section{Discussion}
In this paper, we propose a box-packing strategy with a vision-based high-resolution tactile sensor GelSlim~\cite{GelSlim}. The contributions of the work are: 1) by detecting incipient slip in the packing process with GelSlim, the robot can avoid hard collision with the environment objects. It not only protects the items to pack but also maintains the stability of the grasp and that of the environment objects, enabling future trials. 
2) The image sequences acquired by GelSlim during the contact period are used to estimate both the direction and magnitude of the position error. Based on the estimation, the heuristic control strategy enables the robot to insert new objects within  a few trials. The averaged number of trials for new objects is comparable to that of trained objects. 

Here we briefly comment on the limitations of the proposed strategy and future research directions:
\begin{itemize}
\item The \textbf{control strategy} is based on a hand-crafted heuristic that combines the learned estimations of the error direction and magnitude. It does not take into account the sequence of past actions and observations. Due to the lack of full observability of the contact state between the object and the packed box, a model-based reinforcement learning (RL) framework could come up with a more efficient packing policy with better generalization.
\item The \textbf{dimension} of the error space is small: translation errors in axis $x$ and rotation errors in yaw $\theta$. A more advanced packing policy would also consider tilting the object, which increases the dimension of the action space, but can also provide natural robustness to packing.
\item The \textbf{representation} of the alignment error is a not direct observation, but something that we have to estimate. It would be interesting to compare this method to a purely RL controller that is trained directly on tactile imprints.
\item The \textbf{objects} all have a flat base and all rigid. Exploring generalization to a larger set of objects is important. 
\end{itemize}

The proposed box-packing strategy can be further improved in several ways, but provides a highly flexible baseline that can be applied, for example, on warehouse automation.





\bibliographystyle{IEEEtran}
\bibliography{Ref}

\begin{thebibliography}{10}
\providecommand{\url}[1]{#1}
\csname url@samestyle\endcsname
\providecommand{\newblock}{\relax}
\providecommand{\bibinfo}[2]{#2}
\providecommand{\BIBentrySTDinterwordspacing}{\spaceskip=0pt\relax}
\providecommand{\BIBentryALTinterwordstretchfactor}{4}
\providecommand{\BIBentryALTinterwordspacing}{\spaceskip=\fontdimen2\font plus
\BIBentryALTinterwordstretchfactor\fontdimen3\font minus
  \fontdimen4\font\relax}
\providecommand{\BIBforeignlanguage}[2]{{%
\expandafter\ifx\csname l@#1\endcsname\relax
\typeout{** WARNING: IEEEtran.bst: No hyphenation pattern has been}%
\typeout{** loaded for the language `#1'. Using the pattern for}%
\typeout{** the default language instead.}%
\else
\language=\csname l@#1\endcsname
\fi
#2}}
\providecommand{\BIBdecl}{\relax}
\BIBdecl

\bibitem{correll2018analysis}
N.~Correll, K.~E. Bekris, D.~Berenson, O.~Brock, A.~Causo, K.~Hauser, K.~Okada,
  A.~Rodriguez, J.~M. Romano, and P.~R. Wurman, ``Analysis and observations
  from the first amazon picking challenge,'' \emph{IEEE Transactions on
  Automation Science and Engineering}, vol.~15, no.~1, pp. 172--188, 2018.

\bibitem{GelSlim}
E.~Donlon, S.~Dong, M.~Liu, J.~Li, E.~Adelson, and A.~Rodriguez, ``Gelslim: A
  high-resolution, compact, robust, and calibrated tactile-sensing finger,'' in
  \emph{2018 IEEE/RSJ IROS}.\hskip 1em plus 0.5em minus 0.4em\relax IEEE, 2018,
  pp. 1927--1934.

\bibitem{park2013intuitive}
H.~Park, J.-H. Bae, J.-H. Park, M.-H. Baeg, and J.~Park, ``Intuitive
  peg-in-hole assembly strategy with a compliant manipulator,'' in \emph{IEEE
  ISR 2013}.\hskip 1em plus 0.5em minus 0.4em\relax IEEE, 2013, pp. 1--5.

\bibitem{kim1999active}
I.-W. Kim, D.-J. Lim, and K.-I. Kim, ``Active peg-in-hole of chamferless parts
  using force/moment sensor,'' in \emph{Proceedings 1999 IEEE/RSJ International
  Conference on Intelligent Robots and Systems. Human and Environment Friendly
  Robots with High Intelligence and Emotional Quotients (Cat. No. 99CH36289)},
  vol.~2.\hskip 1em plus 0.5em minus 0.4em\relax IEEE, 1999, pp. 948--953.

\bibitem{drake1978using}
S.~H. Drake, ``Using compliance in lieu of sensory feedback for automatic
  assembly.'' Ph.D. dissertation, Massachusetts Institute of Technology, 1978.

\bibitem{whitney1982quasi}
D.~E. Whitney, ``Quasi-static assembly of compliantly supported rigid parts,''
  \emph{Journal of Dynamic Systems, Measurement, and Control}, vol. 104, no.~1,
  pp. 65--77, 1982.

\bibitem{jain2013scara}
R.~K. Jain, S.~Majumder, and A.~Dutta, ``Scara based peg-in-hole assembly using
  compliant ipmc micro gripper,'' \emph{Robotics and Autonomous Systems},
  vol.~61, no.~3, pp. 297--311, 2013.

\bibitem{bruyninckx1995peg}
H.~Bruyninckx, S.~Dutre, and J.~De~Schutter, ``Peg-on-hole: a model based
  solution to peg and hole alignment,'' in \emph{Proceedings of 1995 IEEE
  International Conference on Robotics and Automation}, vol.~2.\hskip 1em plus
  0.5em minus 0.4em\relax IEEE, 1995, pp. 1919--1924.

\bibitem{newman2001interpretation}
W.~S. Newman, Y.~Zhao, and Y.-H. Pao, ``Interpretation of force and moment
  signals for compliant peg-in-hole assembly,'' in \emph{Proceedings 2001 ICRA.
  IEEE ICRA (Cat. No. 01CH37164)}, vol.~1.\hskip 1em plus 0.5em minus
  0.4em\relax IEEE, 2001, pp. 571--576.

\bibitem{gullapalli1994learning}
V.~Gullapalli, A.~G. Barto, and R.~A. Grupen, ``Learning admittance mappings
  for force-guided assembly,'' in \emph{Proceedings of the 1994 IEEE
  International Conference on Robotics and Automation}.\hskip 1em plus 0.5em
  minus 0.4em\relax IEEE, 1994, pp. 2633--2638.

\bibitem{levine2016end}
S.~Levine, C.~Finn, T.~Darrell, and P.~Abbeel, ``End-to-end training of deep
  visuomotor policies,'' \emph{The Journal of Machine Learning Research},
  vol.~17, no.~1, pp. 1334--1373, 2016.

\bibitem{lee2018making}
M.~A. Lee, Y.~Zhu, K.~Srinivasan, P.~Shah, S.~Savarese, L.~Fei-Fei, A.~Garg,
  and J.~Bohg, ``Making sense of vision and touch: Self-supervised learning of
  multimodal representations for contact-rich tasks,'' \emph{arXiv preprint
  arXiv:1810.10191}, 2018.

\bibitem{chhatpar2001search}
S.~R. Chhatpar and M.~S. Branicky, ``Search strategies for peg-in-hole
  assemblies with position uncertainty,'' in \emph{Intelligent Robots and
  Systems, 2001. Proceedings. 2001 IEEE/RSJ International Conference on},
  vol.~3.\hskip 1em plus 0.5em minus 0.4em\relax IEEE, 2001, pp. 1465--1470.

\bibitem{donahue2015long}
J.~Donahue, L.~Anne~Hendricks, S.~Guadarrama, M.~Rohrbach, S.~Venugopalan,
  K.~Saenko, and T.~Darrell, ``Long-term recurrent convolutional networks for
  visual recognition and description,'' in \emph{Proceedings of the IEEE
  conference on CVPR}, 2015, pp. 2625--2634.

\bibitem{carreira2017quo}
J.~Carreira and A.~Zisserman, ``Quo vadis, action recognition? a new model and
  the kinetics dataset,'' in \emph{proceedings of the IEEE Conference on CVPR},
  2017, pp. 6299--6308.

\bibitem{inception_net}
C.~Szegedy, W.~Liu, Y.~Jia, P.~Sermanet, S.~Reed, D.~Anguelov, D.~Erhan,
  V.~Vanhoucke, and A.~Rabinovich, ``Going deeper with convolutions,'' in
  \emph{Proceedings of the IEEE conference on CVPR}, 2015, pp. 1--9.

\bibitem{nguyen2018translating}
A.~Nguyen, D.~Kanoulas, L.~Muratore, D.~G. Caldwell, and N.~G. Tsagarakis,
  ``Translating videos to commands for robotic manipulation with deep recurrent
  neural networks,'' in \emph{2018 IEEE ICRA}.\hskip 1em plus 0.5em minus
  0.4em\relax IEEE, 2018, pp. 1--9.

\bibitem{finn2016unsupervised}
C.~Finn, I.~Goodfellow, and S.~Levine, ``Unsupervised learning for physical
  interaction through video prediction,'' in \emph{Advances in neural
  information processing systems}, 2016, pp. 64--72.

\bibitem{lee2017learning}
J.~Lee and M.~S. Ryoo, ``Learning robot activities from first-person human
  videos using convolutional future regression,'' in \emph{Proceedings of the
  IEEE Conference on CVPR Workshops}, 2017, pp. 1--2.

\bibitem{yuan2017shape}
W.~Yuan, C.~Zhu, A.~Owens, M.~A. Srinivasan, and E.~H. Adelson,
  ``Shape-independent hardness estimation using deep learning and a gelsight
  tactile sensor,'' in \emph{2017 IEEE ICRA}.\hskip 1em plus 0.5em minus
  0.4em\relax IEEE, 2017, pp. 951--958.

\bibitem{yuan2018active}
W.~Yuan, Y.~Mo, S.~Wang, and E.~H. Adelson, ``Active clothing material
  perception using tactile sensing and deep learning,'' in \emph{2018 IEEE
  ICRA}.\hskip 1em plus 0.5em minus 0.4em\relax IEEE, 2018, pp. 1--8.

\bibitem{li2018slip}
J.~Li, S.~Dong, and E.~Adelson, ``Slip detection with combined tactile and
  visual information,'' in \emph{ICRA}.\hskip 1em plus 0.5em minus 0.4em\relax
  IEEE/RSJ, 2018.

\bibitem{zhang2018fingervision}
Y.~Zhang, Z.~Kan, Y.~A. Tse, Y.~Yang, and M.~Y. Wang, ``Fingervision tactile
  sensor design and slip detection using convolutional lstm network,''
  \emph{arXiv preprint arXiv:1810.02653}, 2018.

\bibitem{ma2018dense}
D.~Ma, E.~Donlon, S.~Dong, and A.~Rodriguez, ``Dense tactile force distribution
  estimation using gelslim and inverse fem,'' in \emph{IEEE ICRA}, 2018.

\bibitem{dong2018slip}
S.~Dong, D.~Ma, E.~Donlon, and A.~Rodriguez, ``Maintaining grasps within
  slipping bound by monitoring incipient slip,'' in \emph{IEEE ICRA}, 2018.

\bibitem{yu2018realtime}
K.-T. Yu and A.~Rodriguez, ``Realtime state estimation with tactile and visual
  sensing for inserting a suction-held object,'' in \emph{2018 IEEE/RSJ
  IROS}.\hskip 1em plus 0.5em minus 0.4em\relax IEEE, 2018, pp. 1628--1635.

\bibitem{Alexnet}
A.~Krizhevsky, I.~Sutskever, and G.~E. Hinton, ``Imagenet classification with
  deep convolutional neural networks,'' in \emph{Advances in neural information
  processing systems}, 2012, pp. 1097--1105.

\bibitem{LSTM}
S.~Hochreiter and J.~Schmidhuber, ``Long short-term memory,'' \emph{Neural
  computation}, vol.~9, no.~8, pp. 1735--1780, 1997.

\bibitem{Zeng2018}
A.~Zeng, S.~Song, K.-T. Yu, E.~Donlon, F.~Hogan, M.~Bauza, D.~Ma, O.~Taylor,
  M.~Liu, E.~Romo, N.~Fazeli, F.~Alet, N.~Chavan-Dafle, R.~Holladay, I.~Morona,
  P.~Q. Nair, D.~Green, I.~Taylor, W.~Liu, T.~Funkhouser, and A.~Rodriguez,
  ``Robotic pick-and-place of novel objects in clutter with multi-affordance
  grasping and cross-domain image matching,'' in \emph{ICRA}.\hskip 1em plus
  0.5em minus 0.4em\relax IEEE, 2018.

\bibitem{kingma2014adam}
D.~Kingma and J.~Ba, ``Adam: A method for stochastic optimization,''
  \emph{arXiv preprint arXiv:1412.6980}, 2014.

\end{thebibliography}

\end{document}